\documentclass[runningheads]{llncs}

\usepackage{eccv}

\usepackage{eccvabbrv}

\usepackage{graphicx}
\usepackage{booktabs}

\usepackage[accsupp]{axessibility}  % Improves PDF readability for those with disabilities.

\usepackage{hyperref}

\hypersetup{
 colorlinks=true,
 linkcolor=red,
 citecolor=cyan,
 filecolor=magenta, 
 urlcolor=cyan,
 }
\usepackage{orcidlink}

\usepackage{orcidlink}

\usepackage{pifont}

\begin{document}

\title{Diffusion-Driven Data Replay: A Novel Approach to Combat Forgetting in Federated Class Continual Learning} 

\titlerunning{Diffusion-Driven Data Replay}

\author{Jinglin Liang$^{1}$\orcidlink{0009-0007-7421-4860} \and
Jin Zhong$^{1}$\orcidlink{0009-0006-5753-2384} \and
Hanlin Gu$^{2}$\orcidlink{0000-0001-8266-4561} \and
Zhongqi Lu$^{3}$\orcidlink{0000-0003-3242-0918} \and
Xingxing Tang$^{2}$\orcidlink{0000-0001-6740-9204} \and
Gang Dai$^{1}$\orcidlink{0000-0001-8864-908X} \and
Shuangping Huang$^{1,5}$\thanks{Corresponding Author}\orcidlink{0000-0002-5544-4544} \and
Lixin Fan$^{4}$\orcidlink{0000-0002-8162-7096} \and
Qiang Yang$^{2,4}$\orcidlink{0000-0001-5059-8360}
}

\authorrunning{J. Liang et al.}

\institute{$^{1}$South China University of Technology, \\
$^{2}$The Hong Kong University of Science and Technology, \\
$^{3}$China University of Petroleum,
$^{4}$WeBank,
$^{5}$Pazhou Laboratory \\
\email{eeljl@mail.scut.edu.cn, eehsp@scut.edu.cn}}

\maketitle

\begin{abstract}
  Federated Class Continual Learning (FCCL) merges the challenges of distributed client learning with the need for seamless adaptation to new classes without forgetting old ones.
  The key challenge in FCCL is catastrophic forgetting, an issue that has been explored to some extent in Continual Learning (CL). However, due to privacy preservation requirements, some conventional methods, such as experience replay, are not directly applicable to FCCL.
  Existing FCCL methods mitigate forgetting by generating historical data through federated training of GANs or data-free knowledge distillation. However, these approaches often suffer from unstable training of generators or low-quality generated data, limiting their guidance for the model.
  To address this challenge, we propose a novel method of data replay based on diffusion models. 
  Instead of training a diffusion model, we employ a pre-trained conditional diffusion model to reverse-engineer each class, searching the corresponding input conditions for each class within the model's input space, significantly reducing computational resources and time consumption while ensuring effective generation. 
  Furthermore, we enhance the classifier's domain generalization ability on generated and real data through contrastive learning, indirectly improving the representational capability of generated data for real data. 
  Comprehensive experiments demonstrate that our method significantly outperforms existing baselines.
  Code is available at \url{https://github.com/jinglin-liang/DDDR}.

  \keywords{Federated Continual Learning \and Diffusion Model}
\end{abstract}

\section{Introduction}
\label{sec:intro}

Federated Learning (FL) \cite{Fedavg,Fedprox} is an emerging machine learning paradigm, offering a way to perform collective learning across decentralized devices while ensuring data privacy. Its ability to protect privacy makes it indispensable in domains such as healthcare \cite{nguyen2022federated} and finance \cite{long2020federated}, which are highly sensitive to data security.
However, real-world applications of FL also encounter challenges, such as the introduction of new data classes by clients over time and the variability of participants within the federation.
This situation has led to the development of Federated Class Continual Learning (FCCL) \cite{criado2022non,babakniya2024data}, a novel concept that necessitates models to incorporate new class information into the model during federated training without compromising the knowledge previously acquired.

The key challenge within FCCL is the issue of catastrophic forgetting \cite{mccloskey1989catastrophic}, where the model loses previously acquired knowledge upon learning new tasks. 
While this problem has been explored to some extent in traditional Continual Learning (CL) \cite{lee2017overcoming}, privacy preservation in FCCL introduces unique constraints that may prevent the direct application of existing strategies. 
For instance, experience replay \cite{rolnick2019experience,liu2021rmm}, a leading approach in CL for mitigating forgetting by retaining and rehearsing data from previous tasks, faces significant hurdles in a federated context. Specifically, in privacy-sensitive environments such as healthcare, the prolonged storage of historical data by users might not be permissible \cite{vizitiu2019towards}. Moreover, the departure of federated participants can result in the loss of their stored data, further complicating data management and continuity.

To circumvent these limitations, the forefront works \cite{qi2022better,zhang2023target,babakniya2024data} in FCCL explore training generators to reproduce data from previous tasks.
Specifically, FedCIL \cite{qi2022better} employs a federated training of an improved version of ACGAN \cite{odena2017conditional} to regenerate historical data. However, the training of GANs is known to be relatively unstable, a problem that becomes even more pronounced in federated settings \cite{rasouli2020fedgan}.
Alternative methods \cite{zhang2023target,babakniya2024data} utilize data-free knowledge distillation techniques \cite{chen2019data} to train generators. 
Nevertheless, such methods tend to generate adversarial samples \cite{goodfellow2014explaining} for classifiers. These adversarial samples, due to their significant divergence from the distribution of real data, offer limited guidance capability for the model.
To encapsulate, while current mainstream FCCL methods focus on mitigating model forgetting through generated replay, they encounter issues such as training instability and the inferior quality of generated data.

Inspired by the training stability and high quality of generated data characteristic of diffusion models in image generation \cite{stable_diffusion,sdxl,gal2022image}, we propose the Diffusion-Driven Data Replay (DDDR), an innovative FCCL framework utilizing diffusion models for data replay.
During the learning of each new task, DDDR employs our proposed Federated Class Inversion technique to extract Class embeddings for each new class.
Specifically, we leverage a pre-trained conditional diffusion model for reverse engineering each class, which entails searching within the model's input space for a conditional embedding capable of guiding the model to generate images of the corresponding class. 
This Class embedding can be regarded as a condensed representation of the class, and by preserving this embedding, we can continuously generate data for the current task in subsequent tasks.
The advantages of this approach are twofold. 
On one hand, leveraging powerful pre-trained diffusion models enables the generation of high-quality images. 
On the other hand, Federated Class Inversion requires optimization and communication of only the class embedding parameter, significantly reducing the computational, communication resources, and training time required compared to training the entire diffusion model. 
Subsequently, we employ the class embeddings derived from previous tasks to replay historical data for the classifier, the target model in continuous learning, to mitigate its forgetting. However, despite the high quality of this data, a certain distributional discrepancy from real data persists, potentially impacting model performance.
To mitigate this, we introduce a contrastive learning constraint in the learning of new tasks, aiming to narrow the feature space gap between generated and real data within the same class. This enhances the classifier's generalization ability across both generated and real domains, indirectly boosting the representational capacity of the generated data for the real data.

Our contributions can be summarized as follows:
\begin{itemize}
\item[1)] We propose DDDR, an innovative FCCL framework. This marks the first application of employing the diffusion model to reproduce data in FCCL, effectively mitigating catastrophic forgetting.
\item[2)] We propose Federated Class Inversion, achieving high-quality data generation in federated settings without consuming excessive additional resources.
\item[3)] By incorporating contrastive learning, we enhance the generalization ability of classifiers across generated and real domains, further strengthening the representational capacity of generated data towards real data.
\item[4)] Comprehensive experiments across various datasets demonstrate that our approach significantly outperforms existing methods, establishing a new state-of-the-art (SOTA) benchmark for FCCL.
\end{itemize}

\section{Related Work}

\subsection{Continual Learning}
Continual Learning (CL) aims to develop machine learning models that can learn from a stream of data over time without forgetting previously acquired knowledge. This field has seen the development of various strategies to mitigate catastrophic forgetting \cite{mccloskey1989catastrophic}, broadly classified into four categories: regularization techniques, experience replay, dynamic architectural methods, and generative replay. 
Regularization techniques \cite{kirkpatrick2017overcoming,lee2017overcoming,zenke2017continual,aljundi2018memory} are designed to prevent the model from significantly altering the weights important for previous tasks while learning new ones.
Experience replay \cite{rolnick2019experience,aljundi2019gradient,prabhu2020gdumb,liu2021rmm} involves storing a subset of previously encountered data and periodically retraining the model on this data alongside new information.
Dynamic architectural methods \cite{rusu2016progressive,mallya2018packnet,yoon2018lifelong,von2020continual} involve modifying the network architecture to accommodate new tasks, thereby preserving previous knowledge while expanding the model's capacity.
Generative replay \cite{wu2018incremental,liu2020generative,van2018generative,wang2021ordisco} leverages generative models to synthesize data for past tasks, which is then used to retrain the model alongside new data.
However, given that the above methods were primarily designed for scenarios involving centralized training, they may not be well-suited for contexts where stringent privacy protection is paramount.

\subsection{Federated Continual Learning}
Federated Continual Learning (FCL) represents a fusion of Federated Learning (FL) and Continual Learning (CL), aimed at addressing the dual challenges of learning continuously from data streams across distributed devices while simultaneously safeguarding privacy and data locality \cite{yang2024federated}. 
Early FCL research \cite{yoon2021federated} selectively activates model parameters associated with the current task through the input of a task ID, which necessitates the explicit notification of the model about the current task's ID, introducing additional complexity in task identification and parameter management. GLFC \cite{dong2022federated} demands clients store historical data, raising storage concerns. Park et al. \cite{park2021tackling} focuses on Federated Incremental Domain Learning scenarios, where the number of classes remains constant while domains incrementally evolve. Federated reconnaissance \cite{hendryx2021federated} emphasizes Few-Shot Learning settings within FCL. Some works \cite{chaudhary2022federated,jiang2021fedspeech} focus on applying Federated Continual Learning to tasks beyond image classification. Ma et al. \cite{ma2022continual} distills the knowledge from old to new models using surrogate datasets. The effectiveness of this approach heavily relies on the similarity between the surrogate dataset and clients' local datasets. SOTA works \cite{qi2022better,zhang2023target,babakniya2024data} in FCCL train generators to replicate historical data. Specifically, FedCIL \cite{qi2022better} introduces an enhanced version of ACGAN \cite{odena2017conditional} for federated training, while 
others \cite{zhang2023target,babakniya2024data} apply Data-Free Knowledge Distillation \cite{chen2019data} in a federated context. 

Building upon the concept of generative replay, our approach uniquely employs diffusion models for generation, achieving unprecedented training stability and superior quality of generated data.

\subsection{Diffusion Models in Federated Learning}
Despite the powerful generative capabilities of diffusion models being applied across various domains, their exploration in FL remains limited.
Initial efforts \cite{tun2023federated,jothiraj2023phoenix} have trained diffusion models within the FL framework. These methods, similar to those for training Generative Adversarial Networks (GANs) in a federated setting \cite{li2022ifl,rasouli2020fedgan,wijesinghe2023pfl}, adhere to the principle that privacy is preserved as long as the training process prevents unauthorized access to or inference of clients' local data. They suggest that generating data that is similar but not identical to a client's local data does not constitute a privacy breach, as long as the generated data does not replicate any specific local data.
Conversely, other studies \cite{yang2023exploring,zhang2023federated,yang2023one} have not directly pursued federated training of diffusion models. Instead, they exploit the pre-trained diffusion models to enable one-shot federated learning (OSFL), where the entire training process necessitates only a single round of communication between clients and the server. Clients guide the server's diffusion model by transmitting non-sensitive information such as image features \cite{yang2023exploring}, image descriptions \cite{zhang2023federated}, or classifiers \cite{yang2023one}, which is then used to generate client-specific data on the server. This generated data is subsequently utilized to train classification models server-side.

Diverging from the above methods, we introduce pre-trained diffusion models into the FCCL framework and propose Federated Class Inversion, thereby achieving high-quality generation with reduced resource consumption.

\section{Problem Formulation}
In this section, we specifically formulate the task setting within the FCCL.
FCCL employs a cooperative learning scheme involving a server and $k$ clients collaboratively engaging in a sequence of $n$ tasks, symbolized as $\{\mathcal{T}_1, \mathcal{T}_2, ..., \mathcal{T}_n\}$.
Each task $\mathcal{T}_t$ corresponds to a distinct dataset $\mathcal{B}_t$, which is distributed among the $k$ clients. This dataset comprises image-label pairs for image classification and is formatted as $\mathcal{B}_t = \{(x_i, y_i) | i = 1,2,...,m\}$. 
Here, $x_i$ represents the images, while $y_i$ denotes their associated class labels, all of which are part of the label set $\mathcal{Y}_t$ for the task $\mathcal{T}_t$.
A fundamental feature of this setup is the mutual exclusivity of label sets across tasks, guaranteed by $\mathcal{Y}_{t1} \cap \mathcal{Y}_{t2} = \emptyset$ for any $t1 \ne t2$. 
Additionally, during the learning of the current task, access to data from previous tasks and the Task ID is restricted.
The objective is to equip the model after learning all tasks with the capability to accurately classify images across the cumulative label set $\bigcup_{t=1}^{n} \mathcal{Y}_t$.

\section{Diffusion-Driven Data Replay}

\begin{figure}[t]
    \centering
    \includegraphics[width=1\textwidth]{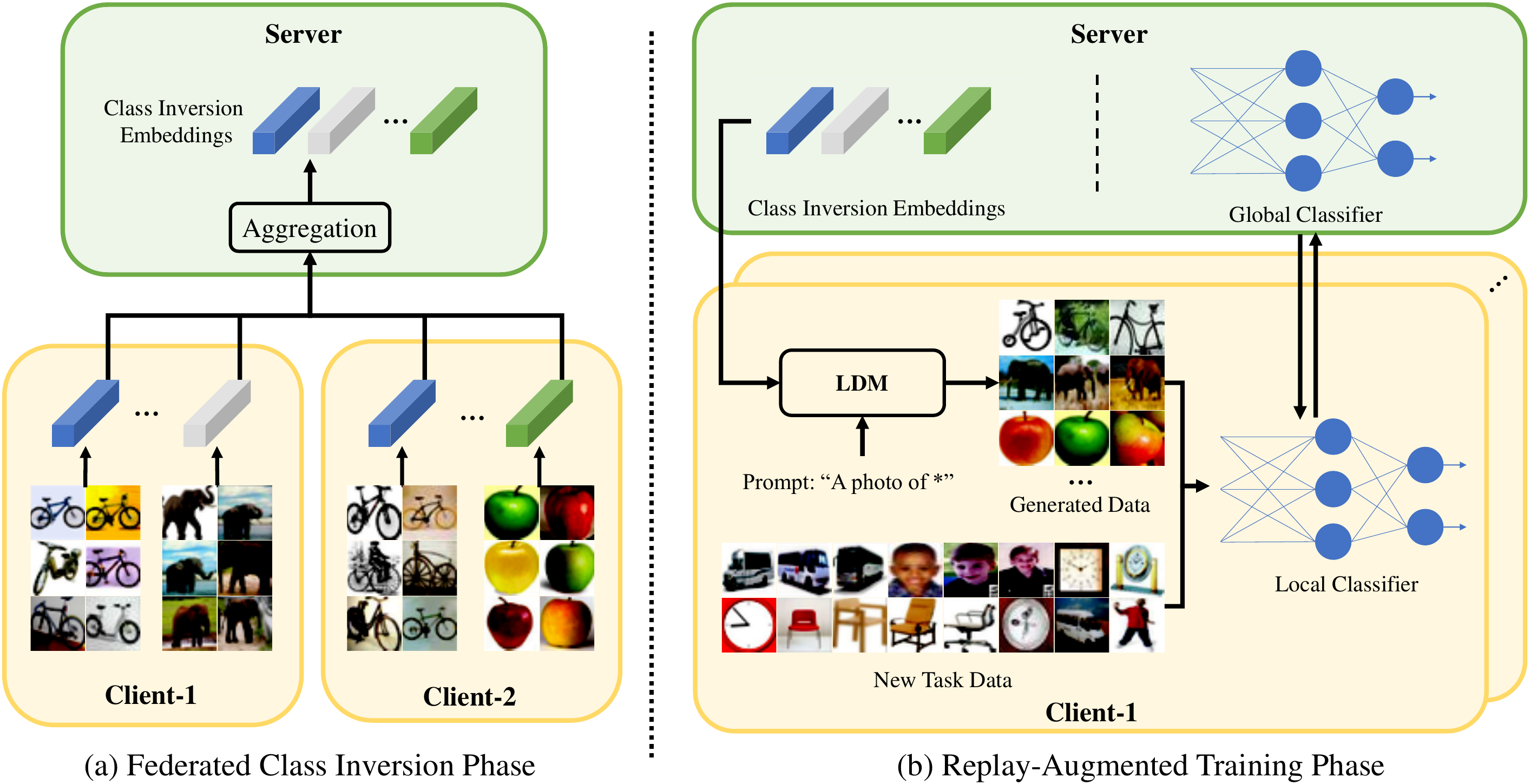}
    \caption{Overview of the DDDR Framework. 
    (a) Federated Class Inversion Phase, in which a pre-trained diffusion model is utilized to reverse-engineer an embedding for each class. This embedding serves as a condensed representation of all images within the class, efficiently encapsulating the essence of the class in a compact vector.
    (b) Replay-Augmented Training Phase, in which clients employ the diffusion model along with previously obtained embeddings to regenerate data. Subsequently, clients train classifiers using the generated data and real data from new tasks.
    % Notably, during classifier training, we introduce contrastive learning loss to unify the model's representation of generated and real data, thereby indirectly bolstering the representativeness of generated data for real scenarios.
    % (a) Federated Class Inversion Phase, where a pre-trained diffusion model is used to reverse engineer embeddings for each class category, serving as a compressed representation of all images within that category. These embeddings are aggregated on the server. (b) Replay-Augmented Training Phase, where clients utilize the diffusion model along with the obtained embeddings from the previous phase to replay historical data for classifier training, facilitating the enhancement of the global classifier with generated data.
    }
    \label{fig-overview}
\end{figure}

As illustrated in Figure \ref{fig-overview}, the DDDR framework encompasses two phases: the Federated Class Inversion Phase and the Replay-Augmented Training Phase. During the first phase, we extract and save an embedding for each class. In the second phase, we utilize the embeddings obtained from the first phase to replay data, and these generated data, along with real data, are used to train the classifier. Detailed descriptions of these phases are provided subsequently.

\subsection{Federated Class Inversion Phase}

Replaying historical data is an effective strategy to counter catastrophic forgetting \cite{rolnick2019experience,aljundi2019gradient}. To achieve this in contexts where data retention is not permissible, an intuitive solution is to train a generator capable of reproducing data from previous tasks. 
The advanced generative capabilities of diffusion models inspire their application in FCCL to generate historical data. However, training a diffusion model for each task is impractical due to the significant time and computational resources required, and the quality of generation may be compromised with limited client data.

To address this challenge, inspired by works in personalized generative models \cite{gal2022image} and image editing \cite{kawar2023imagic}, we propose a novel approach known as Federated Class Inversion.
This method utilizes a frozen, pre-trained diffusion model to conduct reverse engineering on images from various classes, searching for conditional embeddings that can guide the diffusion model to generate images of the corresponding classes. This strategy negates the need for training a diffusion model for each task, substantially reducing the computational burden.

Subsequently, we will introduce Federated Class Inversion from three aspects: the pre-trained Latent Diffusion Model we employ, the local training for Class Inversion, and the global aggregation of Class Embeddings.

\subsubsection{Latent Diffusion Model.}

% We select the Latent Diffusion Model (LDM) to implement our method due to its fast inference speed and the wide availability of open-source model parameters.  Notably, our approach is not limited to using only LDM, other types of pre-trained diffusion models can also be employed within our method.
Theoretically, any pre-trained conditional diffusion model can be used for Federated Class Inversion. In this work, we choose the Latent Diffusion Model (LDM) due to its fast inference speed and widely available pre-trained weights.

The Latent Diffusion Model comprises two primary components: an autoencoder and a diffusion model. The autoencoder \cite{van2017neural,agustsson2017soft} consists of an encoder, $\mathcal{E}$, and a decoder, $\mathcal{D}$. 
% The encoder, $\mathcal{E}$, maps input images $x$ to a spatial latent code, denoted as $z = \mathcal{E}(x)$, while the decoder $\mathcal{D}$ is trained to perform the inverse mapping of the encoder, achieving an approximation $x \approx \mathcal{D}(\mathcal{E}(x))$.
The encoder $\mathcal{E}$ maps the input image $x$ to a low-dimensional latent code $z = \mathcal{E}(x)$ to reduce the computational load of subsequent denoising, while the decoder $\mathcal{D}$ is trained to perform the inverse mapping of the encoder.

The diffusion model, the second component, is tasked with denoising the latent codes.
It is a conditional U-net \cite{ho2022classifier}, where the condition can be derived from various sources such as text, segmentation maps, etc. In our context, we focus solely on textual inputs. 
Its training objective is to predict the noise added to a latent code based on the input condition and a noise-corrupted version of the code. The mathematical formulation of this objective is given by:
\begin{align}
\mathcal{L}_{LDM}=\mathbb{E}_{z\sim\mathcal{E}(x),p,\epsilon\sim\mathcal{N}(0,1),t}\left[\|\epsilon-\epsilon_\theta(\sqrt{\alpha_t}z+\sqrt{1-\alpha_t}\epsilon,t,c_\theta(p))\|_2^2\right],
\end{align}
where $z$ is the latent code of the input image $x$ generated by the encoder $\mathcal{E}$, $\epsilon$ refers to noise sampled from the standard normal distribution $\mathcal{N}(0,1)$, $t$ denotes the timestep in the diffusion process, and $\alpha_t$ is a hyperparameter related to $t$, $p$ represents the text condition, $c_\theta(p)$ denotes the word embedding obtained by encoding $p$ using the text encoder $c_\theta$, and $\epsilon_\theta$ is the model that predicts the noise.

% In this work, we do not train a latent diffusion model, instead, we harness a pre-trained latent diffusion model for image generation. 
% Specifically, we begin by sampling Gaussian noise within the latent space. This noise is then progressively denoised step by step using the diffusion model. Finally, the denoised noise is mapped back to image space using the decoder $\mathcal{D}$.

\subsubsection{Local Class Inversion.}

\begin{figure}[t]
    \centering
    \includegraphics[width=1\textwidth]{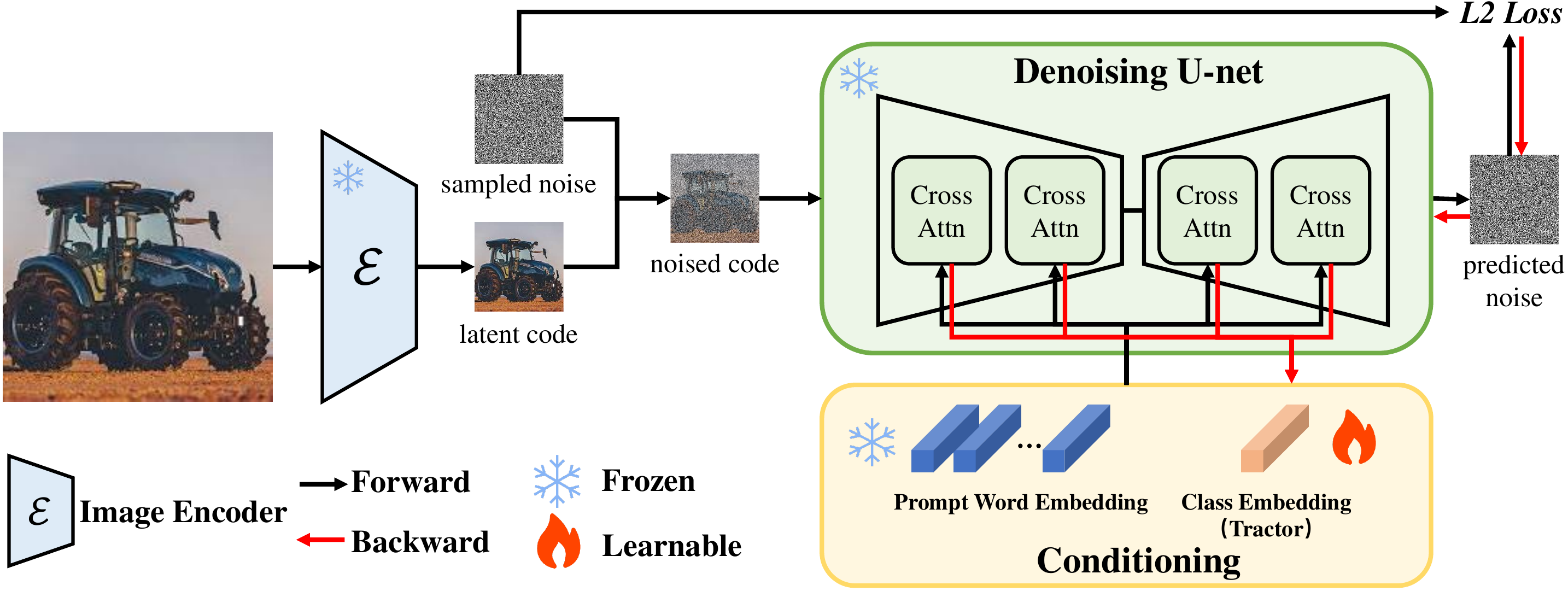}
    \caption{Demonstrating the Local Class Inversion using the tractor class as an example. Initially, a tractor image is sampled from the client's local dataset and fed into the encoder $\mathcal{E}$, yielding the latent code. Concurrently, a frozen prompt's word embedding is concatenated with a learnable class embedding to form a guiding condition. This condition, along with the noise-added latent code, is inputted into the diffusion model to calculate the loss. The class embedding is then optimized using this loss.}
    \label{fig-inversion}
\end{figure}

To capture class embeddings for further replay, we utilize a pre-trained text-to-image LDM for reverse engineering on images from every class, a process termed Class Inversion. Specifically, our objective is to search within the text embedding space, which serves as the input space for the LDM, for an embedding that can instruct the diffusion model to recreate images of a given class. 
To accomplish this, as illustrated in Figure \ref{fig-inversion}, a prompt $p$, such as ``a photo of'', is encoded using a frozen text encoder $c_\theta$ into a word embedding $c_\theta(p)$. A learnable embedding $v$ is then randomly initialized and concatenated with $c_\theta(p)$ to form the guiding condition [$c_\theta(p)$; $v$]. This combined condition is employed to compute the LDM's loss function. Our optimization goal is formulated as:
\begin{align}\label{equal-local-invert}
v^{*}_{i}=\mathop{\arg\min}\limits_{v}\mathbb{E}_{z\sim\mathcal{E}(X_i),p,\epsilon\sim\mathcal{N}(0,1),t}\left[\|\epsilon-\epsilon_\theta(\sqrt{\alpha_t}z+\sqrt{1-\alpha_t}\epsilon,t,[c_\theta(p);v])\|_2^2\right],
\end{align}
where $X_i$ denotes the set of images belonging to the $i^{th}$ class, and $v_i$ represents the class embedding for class $i$.

During the Federated Class Inversion Phase, each client locally optimizes these class embeddings $v_i$ utilizing Equation \ref{equal-local-invert}.
% without the necessity to train the frozen model parameters $c_\theta$ and $\epsilon_\theta$.
This method strategically focuses on optimizing and communicating only the class embeddings with the server, significantly reducing the computational and communication resources required compared to training the entire generative model.
We provide a detailed analysis of time and transmission efficiency in the supplementary materials.

\subsubsection{Global Class Embedding Aggregation.}

In a federated setting, images of each class are distributed across various clients. To train a global class embedding for each class without transmitting data, we employ the FedAvg \cite{Fedavg} algorithm to aggregate the class embeddings. The aggregation is represented as:
\begin{align}
v_{i}=\frac{1}{k}\sum_{j=1}^{k}v_{i}^{(j)},
\end{align}
where $v_{i}^{(j)}$ denotes the class embedding for class $i$ uploaded by the $j$ client, while $v_{i}$ represents the global class embedding for class $i$.

By iteratively performing local training and global aggregation until the class embeddings converge, we obtain the class embeddings for each class. These embeddings are preserved for subsequent data replay.

\subsubsection{Privacy.} 
During the Federated Class Inversion Phase, akin to other distributed training approaches for generative models \cite{tun2023federated,li2022ifl}, our method trains by transmitting learnable parameters rather than disseminating data, inherently providing a layer of privacy protection. 
Moreover, we confine our optimization to the input space of the LDM without modifying the parameters of the pre-trained LDM, thereby not altering its output space. 
Consequently, the probability of generating images that are identical to the original data with our method is lower compared to those approaches that adjust the model's parameters. 
Should a higher level of privacy protection be required, methodologies such as differential privacy \cite{wei2020federated} can be seamlessly incorporated into our framework.
We provide further analysis of privacy issues in the supplementary materials.

\subsection{Replay-Augmented Training Phase}

In this phase, we first utilize the class embeddings obtained during the Federated Class Inversion Phase to guide the diffusion model in generating data. Subsequently, both the generated and real data are employed together to train the classifier.

\subsubsection{Data Generation.} 

To mitigate the catastrophic forgetting of knowledge from previous tasks, we employ the class embeddings learned from previous tasks to generate historical classes images $\hat{\mathcal{X}}_p$ and their corresponding labels $\hat{\mathcal{Y}}_p$.
Additionally, considering the non-IID challenge that can affect model training stability and performance, we also generate current task images $\hat{\mathcal{X}}_c$ and their labels $\hat{\mathcal{Y}}_c$ using the class embeddings of current task classes. 
By ensuring all clients share a similar distribution of generated data, this approach mitigates the extent of non-IID challenges.

\subsubsection{Training Classifier.}

The training process for the classifier aims to fulfill two objectives: learning from 
the current task and revisiting previous tasks.

For learning the new task, we calculate the cross-entropy loss for both the real and generated data of the current task, formulated as:
\begin{align}\label{eq-lce}
\mathcal{L}_{CE}=\mathbb{E}_{x\sim\mathcal{X}_c \cup \hat{\mathcal{X}}_c ,y\sim\mathcal{Y}_c \cup \hat{\mathcal{Y}}_c}[CE(\mathcal{F}(x),y)],
\end{align}
where $\mathcal{F}$ refers to the classifier,  $\mathcal{X}_c$ denotes the real image set of the current task, and $\mathcal{Y}_c$ represents their corresponding labels.

Moreover, since the LDM was not tuned on the clients' data, the generated data may exhibit domain discrepancies from the real data. To address this challenge, we employ a supervised contrastive learning loss \cite{khosla2020supervised,dai2023disentangling} to constrain the classifier's feature space. By narrowing the feature representation gap between generated and real data within the same class, we enhance the model's generalization ability across both the generated data domain and real data. This approach indirectly strengthens the representational capability of the generated data towards real data.
Specifically, we extract features before the final fully connected layer of the classifier, denoted as $e=\mathcal{F}_{e}(x)$, where $\mathcal{F}_{e}$ represents the feature extraction part of the classifier $\mathcal{F}$, and $x$ signifies an individual image. These extracted features are then employed to compute the loss through the following formulation:
\begin{align}\label{eq-lscl}
\mathcal{L}_{SCL}=\mathbb{E}_{e_{i} \sim \mathcal{F}_{e}(\mathcal{X}_c \cup \hat{\mathcal{X}}_c),e_p \sim P(e_i)}[\log\frac{\exp\left(sim\left(e_{i},e_{p}\right)/\tau\right)}{\sum_{i \ne j}\exp\left(sim\left(e_{i},e_{j}\right)/\tau\right)}],
\end{align}
where $P(e_i)$ denotes the set of positive samples that belong to the same class as $e_i$, $sim(e_i,e_j)=f_1(e_i)^{T} f_1(e_j)$ is the similarity function, $f_1$ represents a multilayer perceptron (MLP) that maps features into an $l_2$-normalized feature space, and $\tau$ is a temperature coefficient.

For revisiting the old task, we employ two loss functions. Initially, we compute the cross-entropy loss directly using generated data from old tasks, expressed as:
\begin{align}\label{eq-lpce}
\mathcal{L}_{PCE}=\mathbb{E}_{x\sim\hat{\mathcal{X}}_p ,y\sim\hat{\mathcal{Y}}_p}[CE(\mathcal{F}(x),y)].
\end{align}

Subsequently, to transfer the knowledge from historical tasks to the new model, we employ a knowledge distillation approach \cite{hinton2015distilling} on the generated dataset to migrate the knowledge of the model trained on the previous task to the current model. The knowledge distillation loss function can be represented as:
\begin{align}\label{eq-lkd}
\mathcal{L}_{KD}=\mathbb{E}_{x\sim\hat{\mathcal{X}}_p}[KL(\mathcal{F}(x),\mathcal{F}^{\prime}(x))],
\end{align}
where $KL$ denotes the calculation of the Kullback-Leibler divergence, and $\mathcal{F}^{\prime}$ represents the model preserved after training on the previous task.

In summary, the final objective function for training the classifier on the client side is formulated as:
\begin{align}
\mathcal{F}^{*}=\mathop{\arg\min}\limits_{\mathcal{F}}\mathcal{L}_{CE}+w_{1}\mathcal{L}_{SCL}+w_{2}\mathcal{L}_{PCE}+w_{3}\mathcal{L}_{KD},
\end{align}
where $w_{1}$, $w_{2}$, and $w_{3}$ are hyperparameters used to balance the contributions of each term.

On the server side, we employ the FedAvg \cite{Fedavg} algorithm to aggregate the parameters of the classifiers.

\section{Experiments}

\subsection{Experimental Settings}

\subsubsection{Dataset.}
We conduct experiments on two datasets: Cifar-100 \cite{krizhevsky2009learning} and Tiny-ImageNet \cite{le2015tiny}. To simulate a class-incremental learning scenario, we employ a widely utilized data partitioning paradigm \cite{zenke2017continual,babakniya2024data}. 
Specifically, we divide the dataset classes evenly according to the predetermined task number, with each subset corresponding to the dataset for a particular task. 
% This ensures that the dataset categories for different tasks are mutually exclusive. 
In this work, we set the number of tasks to either 5 or 10. 
Additionally, we consider both IID and non-IID scenarios. For IID, we evenly distribute the data of each class among all clients. For non-IID, we adopt the widely accepted practice \cite{Fedprox,zhang2023target} in FL of using the Dirichlet distribution to simulate an imbalanced label distribution across various clients. In our experiments, the Dirichlet parameter is set to 0.5.

\subsubsection{Evaluation metric.}
We follow the principal works in FCCL \cite{zhang2023target,babakniya2024data}, employing two evaluation metrics: average accuracy and forgetting measure \cite{chaudhry2018riemannian}. Average accuracy is the mean accuracy achieved by the model on all classes after the completion of training across all tasks. The forgetting measure is the mean difference between the peak accuracy and the final accuracy for each class, reflecting the extent to which the model forgets previously learned tasks.

\subsubsection{Compared methods.}
We conducted comparisons with four methods: 
1) \textbf{Finetune}, involving directly fine-tuning on subsequent tasks. 
2) \textbf{FedEWC}, which implements the classical regularization-based continual learning strategy EWC \cite{lee2017overcoming} within the Fedavg \cite{Fedavg} framework. 
3) \textbf{Target} \cite{zhang2023target}, an FCCL work that leverages data-free knowledge distillation for generative replay. 
4) \textbf{MFCL} \cite{babakniya2024data}, a contemporaneous work with Target, follows a similar generative replay strategy as Target. However, it distinguishes itself by designing a more sophisticated loss function for the generator, achieving superior performance. It is the current SOTA in the field of FCCL.

\subsubsection{Implementation detail.}
To ensure a fair comparison, all comparative methods and our approach utilize ResNet-18 \cite{he2016deep} as the classifier, with all experiments conducted using 5 clients. During the Federated Class Inversion Phase, we employ the LDM pre-trained on the LAION-400M dataset \cite{schuhmann2021laion}, as proposed by Rombach et al. \cite{stable_diffusion}. 
The training process consists of 10 communication rounds, with each client performing 50 local training steps per round. 
In the Replay-Augmented Training Phase, the number of communication rounds is increased to 100, with each client conducting training for 5 epochs locally per round. The coefficients for the losses, $w_1$, $w_2$, and $w_3$, are set to 1, 0.5, and 10, respectively.

\subsection{Main Results}

% camera-ready
\begin{table}[t]
\tabcolsep=1.3mm
\centering 
\caption{Results of the comparative experiments on the Cifar-100 dataset. `T' indicates the task number. `Acc' denotes average accuracy, with higher values indicating better performance, and `FM' represents the forgetting measure, where lower values signify lesser forgetting of historical tasks. The best results are highlighted in bold.}
\label{tab-cifar}
\begin{tabular}{c|cccc|cccc}
\toprule
Data partition & \multicolumn{4}{c|}{IID}                                                        & \multicolumn{4}{c}{non-IID}                                                    \\ \midrule
Tasks          & \multicolumn{2}{c|}{T=5}                              & \multicolumn{2}{c|}{T=10} & \multicolumn{2}{c|}{T=5}                            & \multicolumn{2}{c}{T=10} \\ \midrule
Method         & Acc$(\uparrow)$& \multicolumn{1}{c|}{FM$(\downarrow)$}& Acc$(\uparrow)$& FM$(\downarrow)$& Acc$(\uparrow)$& \multicolumn{1}{c|}{FM$(\downarrow)$}& Acc$(\uparrow)$& FM$(\downarrow)$\\ \midrule
Finetune       & 17.33          & \multicolumn{1}{c|}{0.83}            & 9.03        & 0.88        & 16.48          & \multicolumn{1}{c|}{0.81}          & 8.56        & 0.85       \\
FedEWC         & 21.35          & \multicolumn{1}{c|}{0.69}            & 11.76       & 0.73        & 20.96          & \multicolumn{1}{c|}{0.70}          & 11.48       & 0.75       \\
Target         & 34.40          & \multicolumn{1}{c|}{0.48}            & 22.95       & 0.49        & 34.35          & \multicolumn{1}{c|}{0.48}          & 21.71       & 0.51       \\
MFCL           & 42.67          & \multicolumn{1}{c|}{0.37}            & 31.35       & 0.46        & 41.19          & \multicolumn{1}{c|}{0.34}          & 28.99       & 0.41       \\
Ours           & \textbf{51.04} & \multicolumn{1}{c|}{\textbf{0.29}}   & \textbf{43.45}   & \textbf{0.32}   & \textbf{48.45} & \multicolumn{1}{c|}{\textbf{0.26}} & \textbf{41.27}   & \textbf{0.26}  \\ \bottomrule
\end{tabular}
\end{table}

% camera-ready
\begin{table}[t]
\tabcolsep=1.3mm
\centering 
\caption{Results of the comparative experiments on the Tiny-ImageNet dataset.}
\label{tab-tinyimagenet}
\begin{tabular}{c|cccc|cccc}
\toprule
Data partition & \multicolumn{4}{c|}{IID}                                               & \multicolumn{4}{c}{non-IID}                                           \\ \midrule
Tasks          & \multicolumn{2}{c|}{T=5}                   & \multicolumn{2}{c|}{T=10} & \multicolumn{2}{c|}{T=5}                   & \multicolumn{2}{c}{T=10} \\ \midrule
Method         & Acc$(\uparrow)$& \multicolumn{1}{c|}{FM$(\downarrow)$}& Acc$(\uparrow)$& FM$(\downarrow)$& Acc$(\uparrow)$& \multicolumn{1}{c|}{FM$(\downarrow)$}& Acc$(\uparrow)$& FM$(\downarrow)$\\ \midrule
Finetune       & 12.29     & \multicolumn{1}{c|}{0.60}      & 6.80        & 0.67        & 11.68     & \multicolumn{1}{c|}{0.57}      & 6.58        & 0.64       \\
FedEWC         & 13.27     & \multicolumn{1}{c|}{0.49}      & 8.22        & 0.56        & 12.55     & \multicolumn{1}{c|}{0.47}      & 7.66        & 0.52       \\
Target         & 17.56     & \multicolumn{1}{c|}{0.45}      & 12.53       & 0.49        & 17.87     & \multicolumn{1}{c|}{0.41}      & 11.28       & 0.42       \\
MFCL           & 15.11     & \multicolumn{1}{c|}{0.52}      & 10.13       & 0.54       & 13.35     & \multicolumn{1}{c|}{0.48}      & 8.54        & 0.51       \\
Ours           & \textbf{25.47} & \multicolumn{1}{c|}{\textbf{0.36}} & \textbf{19.01}   & \textbf{0.36}   & \textbf{23.96} & \multicolumn{1}{c|}{\textbf{0.33}} & \textbf{16.65}   & \textbf{0.27}  \\ \bottomrule
\end{tabular}
\end{table}

\begin{figure}[t]
    \centering
    \includegraphics[width=0.85\textwidth]{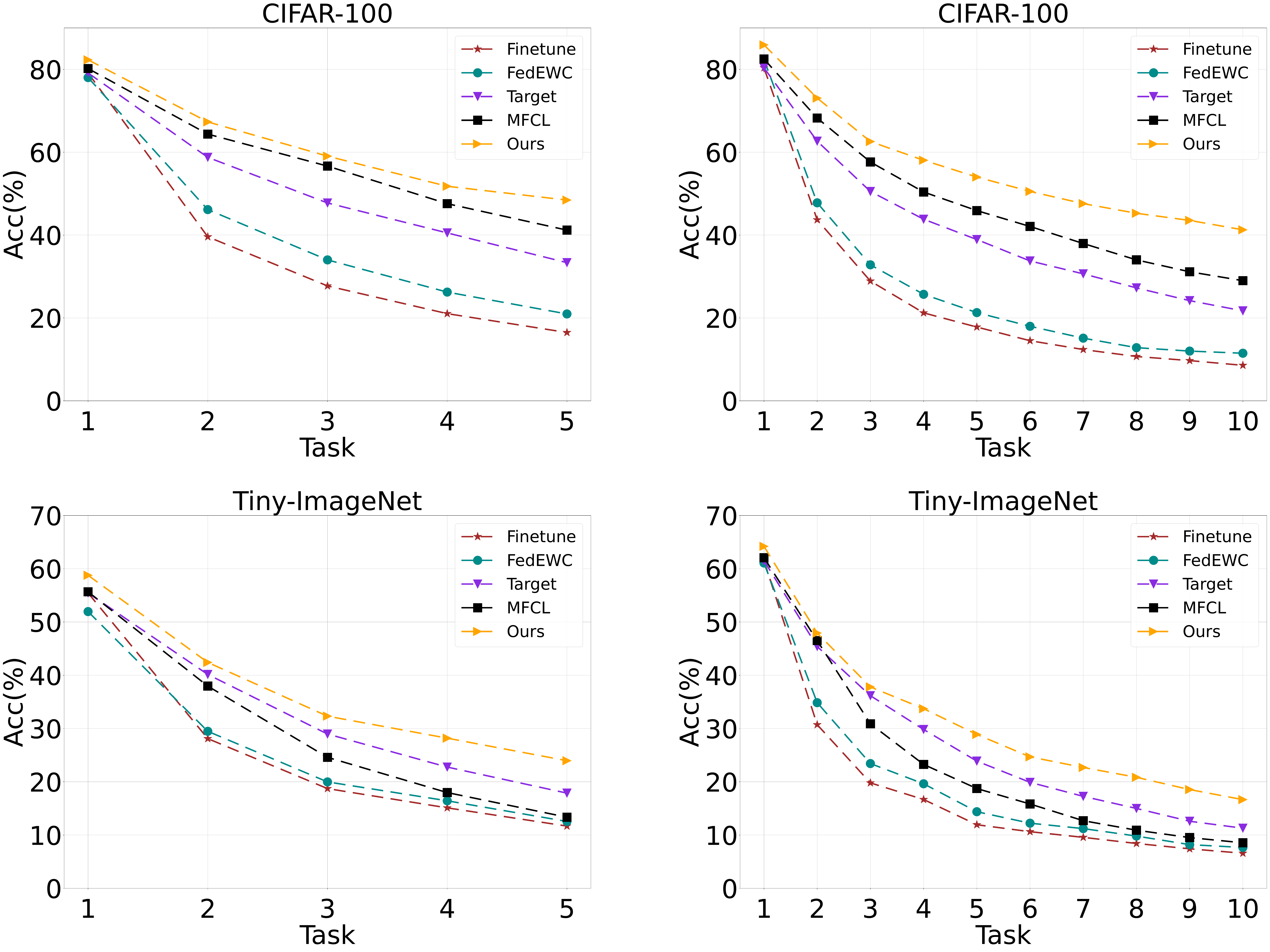}
    \caption{Details of the variation in average accuracy as the learned task number increases across different methods on the CIFAR-100 and TinyImageNet datasets with a non-IID distribution of data.}
    \label{fig-accchart}
\end{figure}

\begin{table}[t]
\tabcolsep=2.4mm
\centering
\caption{Results of ablation experiments for DDDR on the Cifar-100 dataset. The number of tasks is set to 5, with a non-IID distribution of data. $(\hat{\mathcal{X}}_p, \hat{\mathcal{Y}}_p)$ and $(\hat{\mathcal{X}}_c, \hat{\mathcal{Y}}_c)$ respectively denotes the generated data for historical tasks and current task. $\mathcal{L}_{SCL}$ signifies the contrastive learning loss function. The symbols \ding{51} and \ding{55} respectively indicate the inclusion and ablation of the corresponding settings.}
\label{tab-ablation}
\begin{tabular}{c|ccc|cc}
\toprule
  & $(\hat{\mathcal{X}}_p, \hat{\mathcal{Y}}_p)$ & $(\hat{\mathcal{X}}_c, \hat{\mathcal{Y}}_c)$ & $\mathcal{L}_{SCL}$ & Acc$(\uparrow)$ & FM$(\downarrow)$   \\ \midrule
1  & \ding{51}                       & \ding{51}                       & \ding{51}                        & 48.45                    & 0.26 \\
2  & \ding{55}                       & \ding{51}                       & \ding{51}                        & 17.63                    & 0.84 \\
3  & \ding{51}                       & \ding{55}                       & \ding{51}                        & 44.29                    & 0.36 \\
4  & \ding{51}                       & \ding{51}                       & \ding{55}                        & 45.34                    & 0.28 \\
5  & \ding{55}                       & \ding{55}                       & \ding{51}                        & 16.37                    & 0.82 \\
6  & \ding{51}                       & \ding{55}                       & \ding{55}                        & 45.13                    & 0.29 \\
7  & \ding{55}                       & \ding{51}                       & \ding{55}                        & 17.51                    & 0.83 \\
8  & \ding{55}                       & \ding{55}                       & \ding{55}                        & 16.48                    & 0.81 \\ \bottomrule
\end{tabular}
\end{table}

To validate the effectiveness of DDDR, we conducted comparative analyses between DDDR and four existing methods, with the results, averaged over three experiments, presented in Tables \ref{tab-cifar} and \ref{tab-tinyimagenet}, and Figure \ref{fig-accchart}.
From Tables \ref{tab-cifar} and \ref{tab-tinyimagenet}, we can derive the following insights:
1) DDDR demonstrates improved performance over existing methods in all experimental settings across both datasets, in terms of average accuracy and forgetting measure. This suggests that our method effectively mitigates the issue of catastrophic forgetting in FCCL through high-quality data replay, thereby establishing a new SOTA for FCCL.
% 1) DDDR significantly exceeds existing methods in all experimental settings across both datasets, in terms of average accuracy and forgetting measure. This demonstrates that our method can substantially mitigate the issue of catastrophic forgetting in FCCL through high-quality data replay.
2) Finetune yields the poorest results, with its forgetting measure values indicating almost complete forgetting of historical tasks.
3) The regularization-based method FedEWC shows some improvement over Finetune but falls short when compared to generative replay-based methods like Target and MFCL. This suggests that while regularization-based methods can alleviate forgetting to some extent, their effectiveness is limited due to the lack of data-level guidance for the model.
4) The performance of the two generative replay-based methods, Target and MFCL, surpasses the other two baseline methods, yet there remains a discernible gap between them and our DDDR. This disparity is attributed to the lower quality of generated data, which constrains their performance.

Figure \ref{fig-accchart} depicts the changes in average accuracy as the model sequentially learns a series of tasks. We observe a monotonic decline in the average accuracy of all methods as the model progressively learns more tasks, attributed to the forgetting of knowledge from old tasks. Notably, DDDR exhibits the least pronounced decrease in average accuracy across all settings, demonstrating its superior capability to mitigate the forgetting of historical tasks. Moreover, for the initial task, where no historical data exists, our method also achieves the highest accuracy. This indicates that DDDR effectively reduces the impact of non-IID distributions by generating data for the current task, thus improving performance.

\subsection{Ablation Study}
To evaluate the effectiveness of each component of DDDR, we performed ablation experiments, with the results detailed in Table \ref{tab-ablation}. These experiments targeted three main components: generated data for historical tasks $(\hat{\mathcal{X}}_p, \hat{\mathcal{Y}}_p)$, generated data for the current task $(\hat{\mathcal{X}}_c, \hat{\mathcal{Y}}_c)$, and the contrastive learning loss $\mathcal{L}_{SCL}$. 
For the ablation of $(\hat{\mathcal{X}}_p, \hat{\mathcal{Y}}_p)$, we omitted the calculation of the historical task cross-entropy loss from Eq.(\ref{eq-lpce}) and the knowledge distillation loss from Eq.(\ref{eq-lkd}) during classifier training. 
In the case of ablation for $(\hat{\mathcal{X}}_c, \hat{\mathcal{Y}}_c)$, only real data from the current task were used to compute the cross-entropy loss from Eq.(\ref{eq-lce}) and the contrastive learning loss from Eq.(\ref{eq-lscl}), excluding any generated data. 
Finally, for the ablation of the $\mathcal{L}_{SCL}$, we did not calculate the loss specified in Eq.(\ref{eq-lscl}).
Subsequently, we elucidate the role of each component individually. 

\textbf{The Impact of Generated Data for Historical Tasks.} As illustrated in rows 2, 5, 7, and 8 of Table \ref{tab-ablation}, the absence of generated data for historical tasks leads to a significant degradation in both average accuracy and the forgetting measure, with a particularly notable increase in the forgetting measure indicating that the model has almost completely forgotten the historical tasks. This highlights that DDDR's capability to alleviate catastrophic forgetting can be attributed to its ability to generate high-quality data for historical tasks.

\textbf{The Impact of Generated Data for Current Task.} 
% The comparative analysis of rows 1 and 3, alongside rows 4 and 6 in Table \ref{tab-ablation}, illustrates that excluding generated data for the current task reduces both average accuracy and the forgetting measure. These comparisons underscore DDDR's efficacy in ensuring all clients share a similar distribution of generated data, effectively countering the non-IID challenges.
The comparative analysis of rows 1 and 3, alongside rows 4 and 6 in Table \ref{tab-ablation}, demonstrates that excluding generated data for the current task not only reduces the average accuracy but also increases the forgetting measure scores. This indicates the effectiveness of the strategy, which by ensuring all clients share a similarly distributed set of generated data, effectively mitigates the degree of non-IIDness in data distribution, thereby enhancing the model's performance.

\textbf{The Impact of Contrastive Learning Loss.} 
The observations from rows 1 and 4 in Table \ref{tab-ablation} demonstrate that the combined use of $\mathcal{L}_{SCL}$ and $(\hat{\mathcal{X}}_c, \hat{\mathcal{Y}}_c)$ enhances both the average accuracy and the forgetting measure. However, a contrasting review of rows 3, 6, and subsequently rows 5 and 8 reveals that applying $\mathcal{L}_{SCL}$ in the absence of $(\hat{\mathcal{X}}_c, \hat{\mathcal{Y}}_c)$ leads to a deterioration in both average accuracy and the forgetting metric. This indicates that the efficacy of $\mathcal{L}_{SCL}$ is not absolute. It yields significant performance gains when generated data for the current task is included in training, but surprisingly produces adverse effects when such data is omitted. This observation suggests that the performance improvements attributed to $\mathcal{L}_{SCL}$ stem from its ability to bridge the feature representation gap between generated and real data, thereby enhancing the classifier's generalization capability across the generated and real data domains.

\subsection{Visualization of Generated Results}

\begin{figure}[t]
    \centering
    \includegraphics[width=0.91\textwidth]{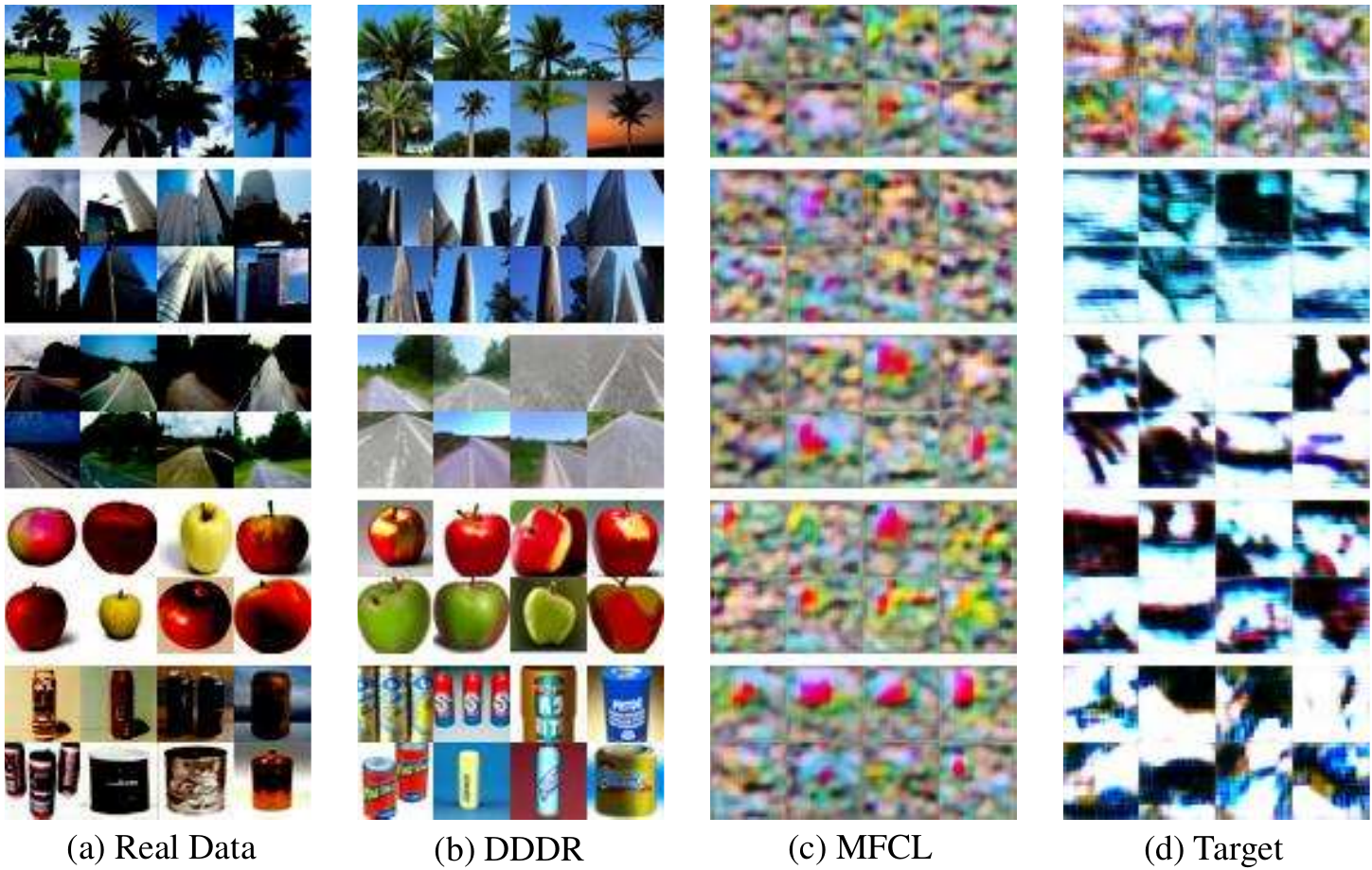}
    \caption{Visualization of generated outcomes from three generative replay methods and the real data from the CIFAR-100 dataset.}
    \label{fig-visualization}
\end{figure}

To analyze the generative capabilities of DDDR's Federated Class Inversion method, we conduct a visual analysis, with the results depicted in Figure \ref{fig-visualization} and supplementary material. 
It is evident that the generated images by DDDR closely approximate the data distribution of real images. 
This indicates that our Federated Class Inversion method can deliver excellent generative results while only requiring a reduced number of optimization parameters.
In contrast, the other two methods based on data-free knowledge distillation, namely Target and MFCL, aim to generate images that mislead the classifier into categorizing them as a specific class. As a result, their generated samples often resemble adversarial examples \cite{goodfellow2014explaining} for the classifier, exhibiting a significant distribution gap from the real data. It can be inferred that the superior generation quality of DDDR is key to more effectively addressing catastrophic forgetting.

\section{Conclusion}
This work proposes DDDR, leveraging diffusion models to replay historical data and address the issue of catastrophic forgetting in FCCL. Compared to existing generative replay methods, our approach is capable of producing higher-quality data. Furthermore, we enhance the domain generalization ability of the classifier on both generated and real data, enabling more efficient utilization of generated data. Comprehensive experiments demonstrate that DDDR significantly outperforms existing approaches, establishing a new state-of-the-art in the FCCL domain.

% \clearpage  % TODO REVIEW/FINAL: This \clearpage needs to be removed from both review and camera-ready versions.

\noindent\textbf{Acknowledgement}
The research is partially supported by National Key Research and Development Program of China (2023YFC3502900), National Natural Science Foundation of China (No.62176093, 61673182), Key Realm Research and Development Program of Guangzhou (No.202206030001), Guangdong-Hong Kong-Macao Joint Innovation Project (No.2023A0505030016).

\bibliographystyle{splncs04}
\bibliography{main}

\clearpage

\title{Supplementary Material} 

\titlerunning{Diffusion-Driven Data Replay}

\author{Jinglin Liang$^{1}$\orcidlink{0009-0007-7421-4860} \and
Jin Zhong$^{1}$\orcidlink{0009-0006-5753-2384} \and
Hanlin Gu$^{2}$\orcidlink{0000-0001-8266-4561} \and
Zhongqi Lu$^{3}$\orcidlink{0000-0003-3242-0918} \and
Xingxing Tang$^{2}$\orcidlink{0000-0001-6740-9204} \and
Gang Dai$^{1}$\orcidlink{0000-0001-8864-908X} \and
Shuangping Huang$^{1,5}$\thanks{Corresponding Author}\orcidlink{0000-0002-5544-4544} \and
Lixin Fan$^{4}$\orcidlink{0000-0002-8162-7096} \and
Qiang Yang$^{2,4}$\orcidlink{0000-0001-5059-8360}
}

\authorrunning{J. Liang et al.}

\institute{$^{1}$South China University of Technology, \\
$^{2}$The Hong Kong University of Science and Technology, \\
$^{3}$China University of Petroleum,
$^{4}$WeBank,
$^{5}$Pazhou Laboratory \\
\email{eeljl@mail.scut.edu.cn, eehsp@scut.edu.cn}}

\maketitle

\setcounter{section}{0}
\renewcommand\thesection{\Alph{section}}
\label{sec:rationale}

We organize the supplementary material as follows.

\begin{itemize}
\item In Section \ref{sec_privacy}, we analyze the privacy protection capabilities of our proposed DDDR framework.
\item In Section \ref{sec_time}, we discuss the time and transmission efficiency of DDDR.
\item In Section \ref{sec_visualization}, we present additional generated samples.
\item In Section \ref{sec_generalizability}, we analyze the generalization capabilities of our proposed Federated Class Inversion. 
\item In Section \ref{sec_local_test}, we present the experimental results of local testing on each client.
\end{itemize}

\section{Privacy concerns}\label{sec_privacy}

\subsection{Integration of privacy protection methods}

\begin{figure}[h]
    \centering
    \includegraphics[width=0.9\textwidth]{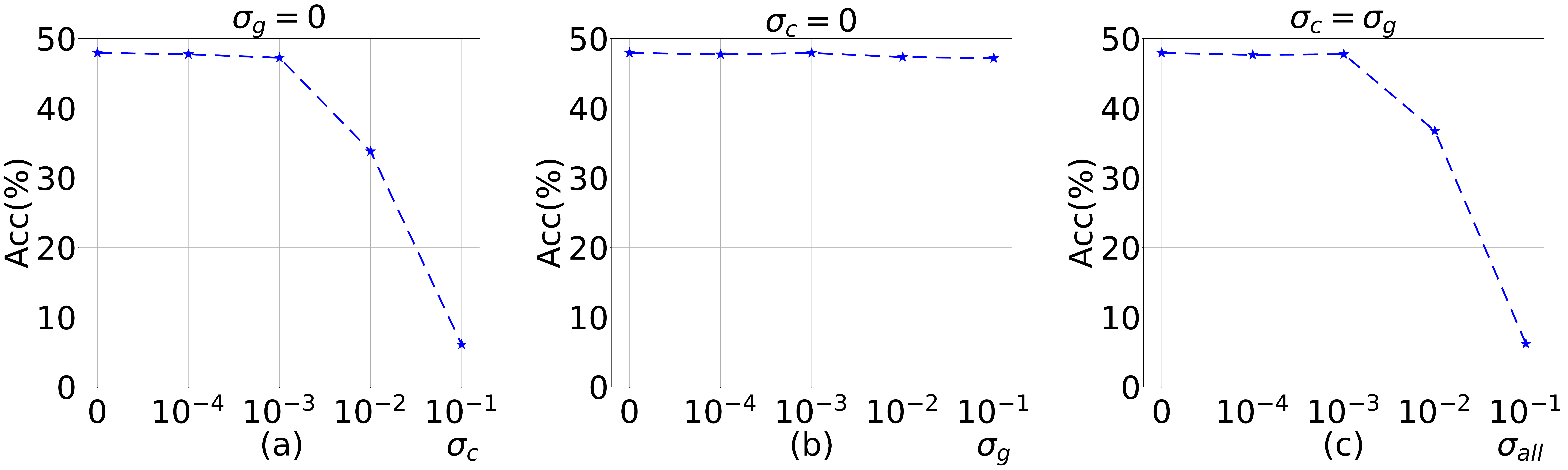}
    \caption{Variations in the average accuracy of DDDR across different noise intensities, on the Cifar-100 dataset with 5 tasks and non-IID data distribution.
    $\sigma_c$ and $\sigma_g$ denote the standard deviations of Gaussian noise introduced to classifier parameters and class embeddings, respectively.
    (a) With $\sigma_g$ set to 0, observing the effect of $\sigma_c$ on average accuracy. (b) With $\sigma_c$ set to 0, examining the impact of $\sigma_g$ on average accuracy. (c) Introducing noise to both class embeddings and classifier parameters to assess their collective influence on average accuracy.}
    \label{fig-dp-accchar}
\end{figure}

To assess the efficacy of privacy protection strategies within the DDDR framework, we incorporate the widely used randomization privacy protection strategy \cite{zhu2019deep,kang2023optimizing} into DDDR. Specifically, during each round of communication, clients first augment their class embeddings and classifier parameters with Gaussian noise before uploading to the server. This approach significantly lowers the success rate of gradient inversion attacks \cite{zhu2019deep}, thus preventing the server or any other federated participants from deducing private data.

Figure \ref{fig-dp-accchar} illustrates the variation in the model's average accuracy with the introduction of noise intensity. As expected, an increase in the noise intensity added to the classifier parameters leads to a reduction in classifier performance, due to the trade-off between privacy protection and model performance \cite{kang2023optimizing}. Unexpectedly, the intensity of noise added to the class embeddings has a minimal impact on model performance. 

\begin{figure}[h]
    \centering
    \includegraphics[width=0.95\textwidth]{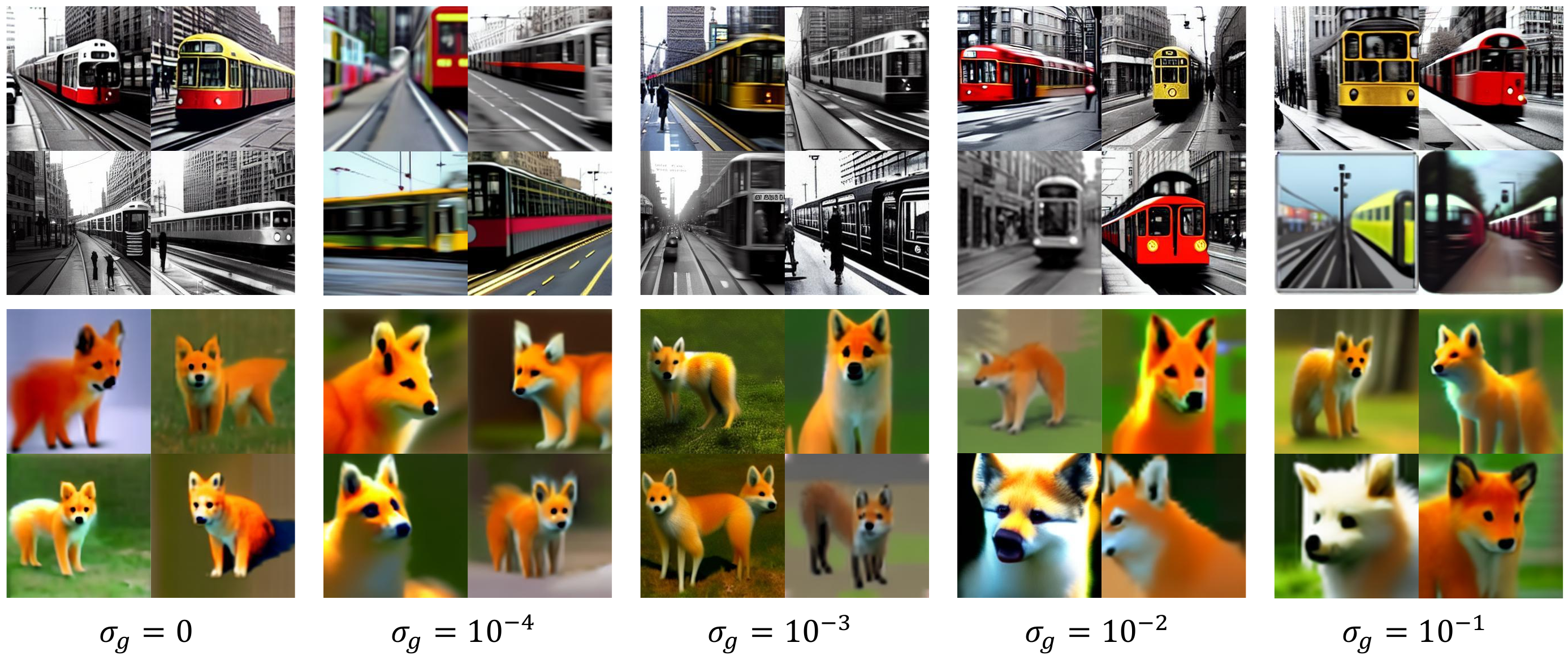}
    \caption{Showcase of DDDR-generated samples under different noise intensities. $\sigma_g$ denotes the standard deviation of noise added to the class embeddings uploaded by clients.}
    \label{fig-visualization-sigma}
\end{figure}

To explore the reasons behind this, we generate images using class embeddings trained under different noise intensities, which are presented in Figure \ref{fig-visualization-sigma}. We observe that the generative quality of class embeddings trained under various noise intensities remains similar. Even at a noise intensity with a standard deviation of 0.1, it is still able to achieve desirable generative outcomes. This may be attributed to the training objective of Federated Class Inversion, which involves searching for an optimal embedding within the input space of a pre-trained conditional diffusion model. Given that this model has been pre-trained on a vast amount of data, its input embedding space is relatively smooth, meaning that perturbations to the embedding do not significantly alter the generative results.

In summary, the randomization privacy protection strategy can be applied to the DDDR framework to enhance privacy protection. Furthermore, our proposed Federated Class Inversion method's generative quality is insensitive to the intensity of noise added, implying that this method can enhance privacy protection without noticeably compromising performance, thereby achieving a effective balance between model performance and privacy protection.

\subsection{Gradient inversion attacks}

\begin{figure}[h]
    \centering
    \includegraphics[width=0.95\textwidth]{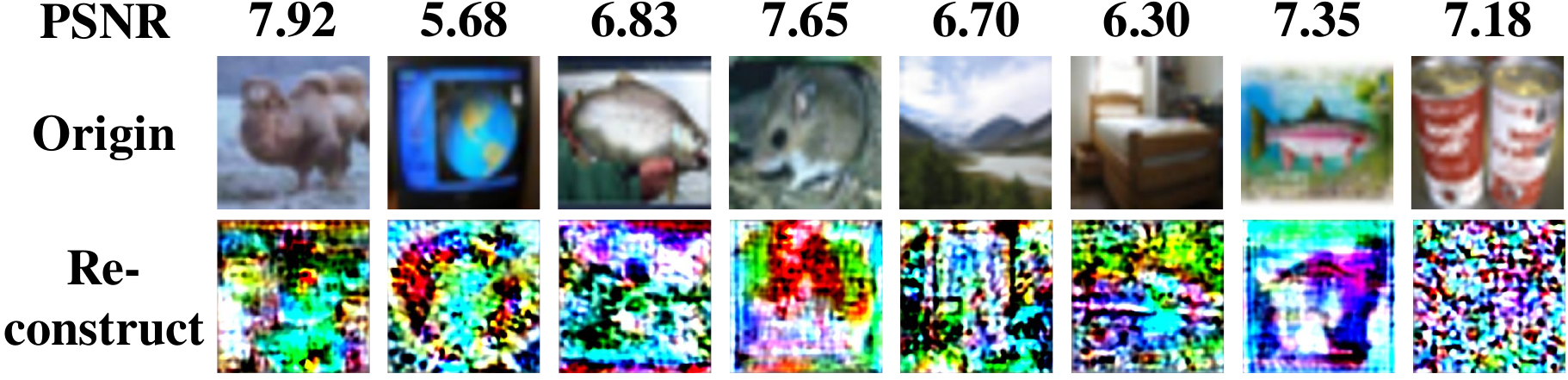}
    \caption{Results of applying gradient inversion attacks on Federated Class Inversion.}
    \label{fig-dgl}
\end{figure}

Transmitting only class embeddings in Federated Class Inversion is secure. We attempted to reconstruct training images from gradients of class embeddings using gradient inversion attacks \cite{zhu2019deep} but were unsuccessful, as shown in the Figure \ref{fig-dgl}.

\subsection{The likelihood of generating the original data}

\begin{figure}[h]
    \centering
    \includegraphics[width=0.95\textwidth]{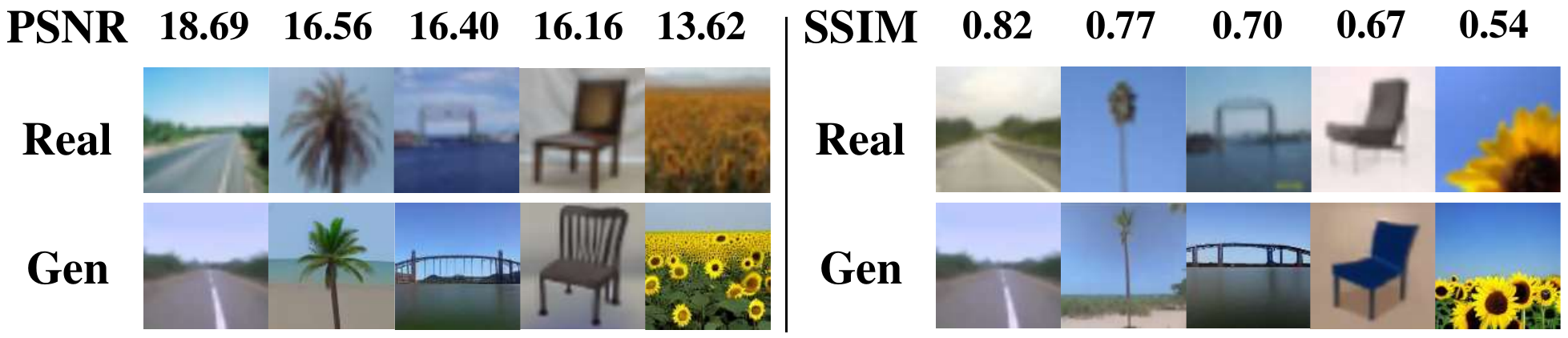}
    \caption{The most similar real-generated image pairs.}
    \label{fig-similarity}
\end{figure}

It is unlikely to generate images that are identical to the original data. 
We randomly selected 5 classes from CIFAR-100 and presented the most similar real-generated image pairs with the highest PSNR or SSIM in the Figure \ref{fig-similarity}. It can be seen that there are noticeable differences between them.

\section{Time and transmission efficiency}\label{sec_time}

To assess whether DDDR's performance improvement comes at the cost of training efficiency, we conduct an analysis of its training time.
In learning a new task, DDDR operates in two stages: the Federated Class Inversion phase, during which a class embedding is optimized for each new category, followed by the Replay-Augmented Training phase, which involves image generation before classifier training. 
Image generation allows for server-side execution without utilizing client computational resources, and the generated images can be stored for repeated use. 
Consequently, Federated Class Inversion and Classifier Training are the two primary factors affecting training time. As shown in Table \ref{tab-time}, for the learning of each new task, the time consumed by Federated Class Inversion is significantly less than that required for classifier training, accounting for only about 12\% of their combined total. 
Comparatively, the training duration for classifiers in DDDR and other baseline methods is similar, given the identical training steps among them, with the primary difference being in the loss function used, which does not significantly impact training time. Thus, the additional time incurred by our method compared to other baselines is attributed to the Federated Class Inversion phase, which is significantly shorter than the time for classifier training and does not substantially affect the overall runtime.

\begin{table}[h]
\tabcolsep=2.4mm
\centering
\caption{Training Time Analysis of DDDR on the Cifar-100 Dataset with Five Tasks. FCI denotes the Federated Class Inversion Phase, CT represents the Classifer Training, and IG stands for Image Generation. The local training duration for one client is reported in minutes for both the FCI and CT phases. For IG, the time required to generate 200 images for a single class is reported. All experiments were conducted on a single 3090 GPU.}
\label{tab-time}
\begin{tabular}{cccc}
\toprule
                    & FCI             & CT                 & IG \\ \midrule
training time (min) & $\approx 4.8$   & $\approx 34.2$     & $\approx 3.63$    \\
communication rounds & 10              & 100                &  -               \\ \bottomrule
\end{tabular}
\end{table}

Moreover, the Federated Class Inversion in DDDR is transmission-efficient, as it only transmits low-dimensional class embeddings. For instance, the transmitting parameter of FCI is at most $128K$ for the diffusion model ($1.5B$) on CIFAR-100.

\section{Visualization of generated results}\label{sec_visualization}

\begin{figure}[htbp]
    \centering
    \includegraphics[width=1\textwidth]{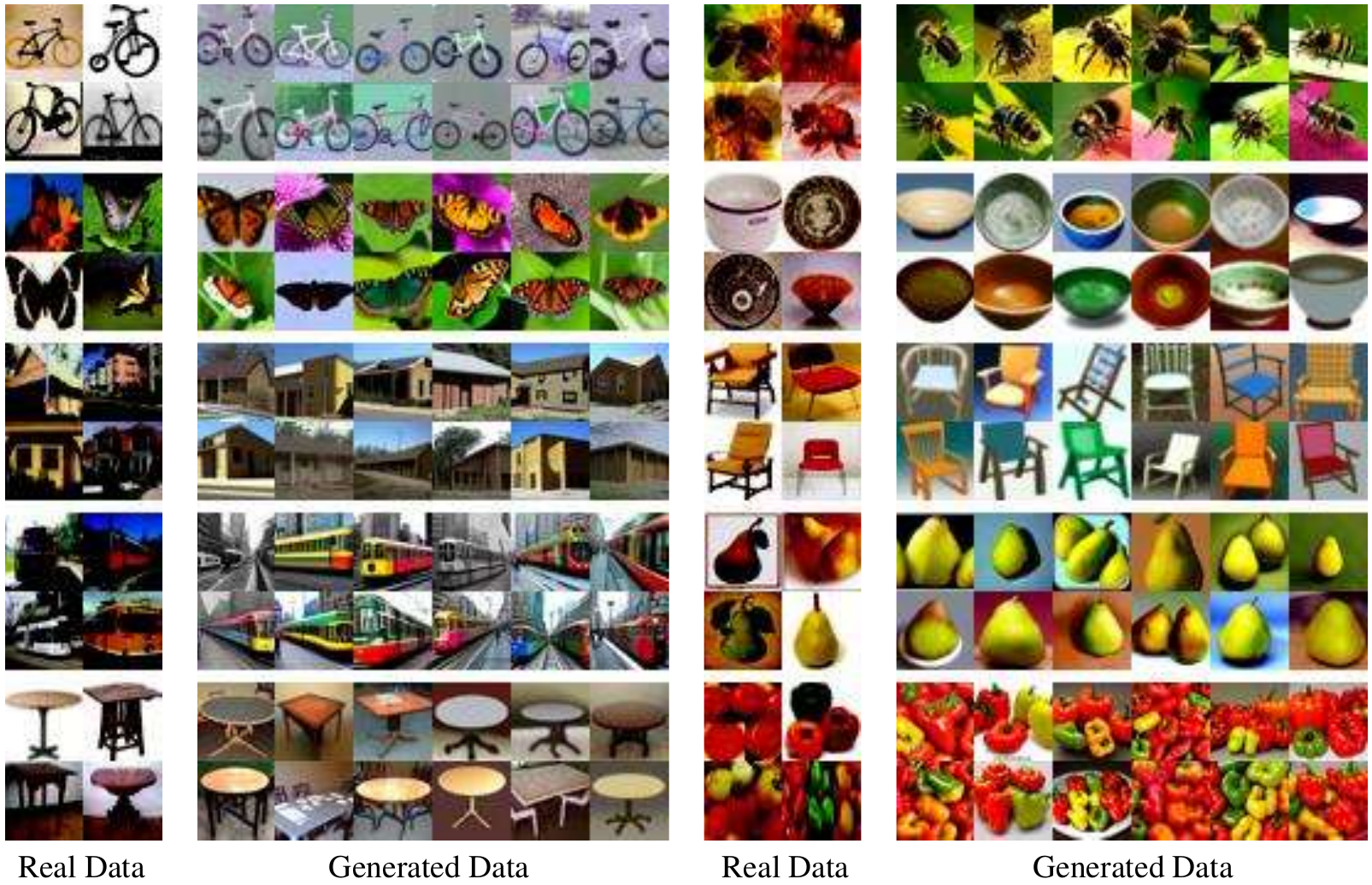}
    \caption{Visualization of generated outcomes from DDDR and the real data from the CIFAR-100 dataset.}
    \label{fig-visualization-cifar100}
    
    \vspace{4em} % Add some vertical space between the images
    
    \includegraphics[width=1\textwidth]{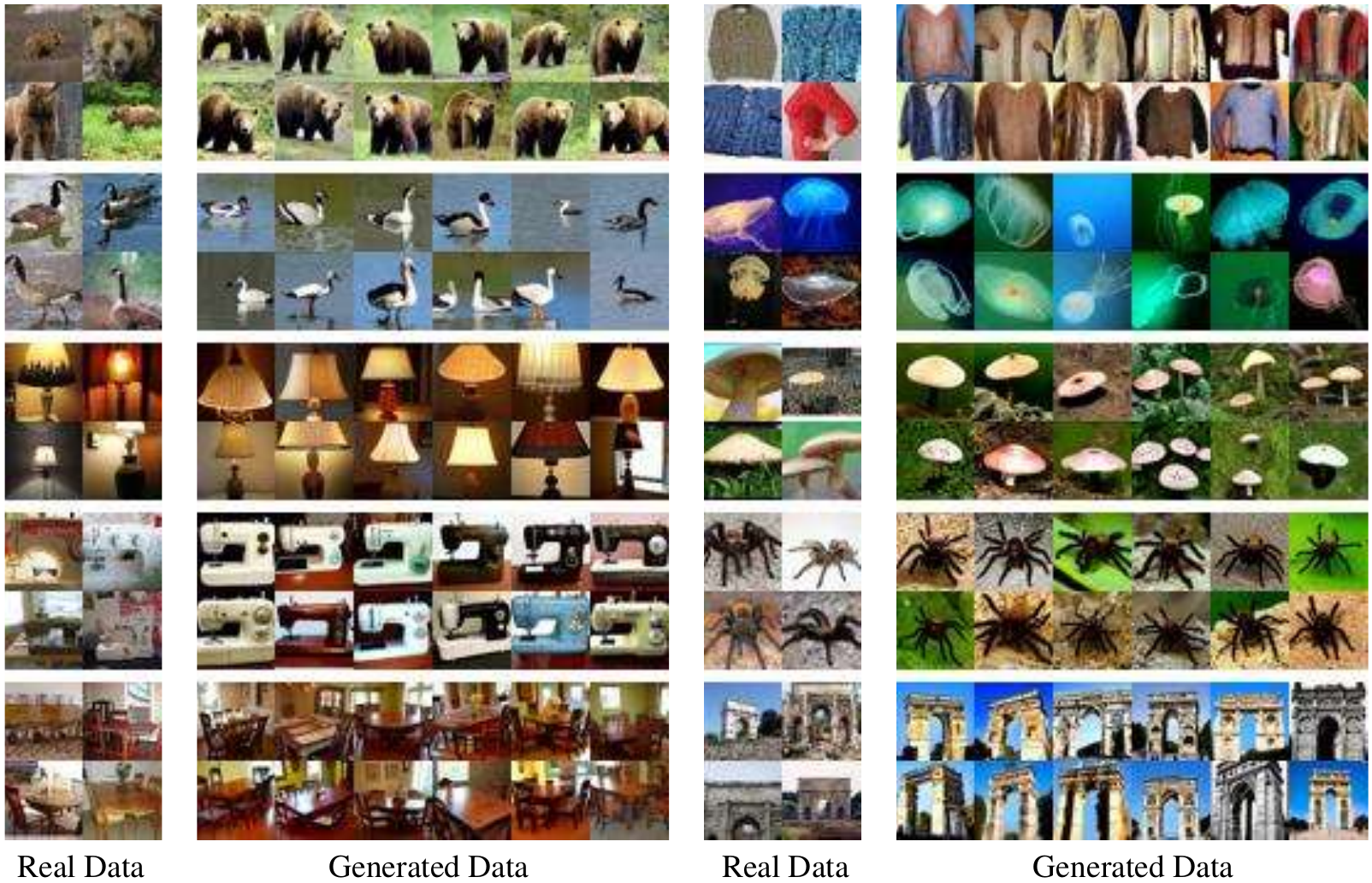}
    \caption{Visualization of generated outcomes from DDDR and the real data from the Tiny-ImageNet dataset.}
    \label{fig-visualization-tinyimagenet}
\end{figure}

To more comprehensively demonstrate the generative capabilities of DDDR, we conduct training for Federated Class Inversion on both the Cifar-100 \cite{krizhevsky2009learning} and Tiny-ImageNet \cite{le2015tiny} datasets. Utilizing the resultant class embeddings, we generate images, with the outcomes presented in Figure \ref{fig-visualization-cifar100} and \ref{fig-visualization-tinyimagenet}.
From the generated results, two observations can be made: 
1) DDDR is capable of producing high-quality images, closely matching the distribution of real images in both diversity and fidelity. For instance, the generated images of categories such as bowls, chairs, and tables in Figure \ref{fig-visualization-cifar100} are highly realistic and exhibit a wide variety of styles and poses. 
2) Despite the high quality of generation, a slight domain discrepancy between the generated and real data is observable \cite{qi2022better,dai2023disentangling}. For example, in Figure \ref{fig-visualization-cifar100}, categories such as buses and houses are more frequently depicted in nighttime scenes in the real data, whereas the generated data tends to favor daytime scenes. This underscores the importance of enhancing the classifier's generalization capability across both the generated and real domains. The cause of this domain discrepancy may be attributed to the limited optimization parameters. In DDDR, to enhance training efficiency, the optimization was conducted solely on the class embeddings without fine-tuning the pre-trained diffusion model.

\section{Generalizability}\label{sec_generalizability}

\begin{figure}[h]
    \centering
    \includegraphics[width=0.85\textwidth]{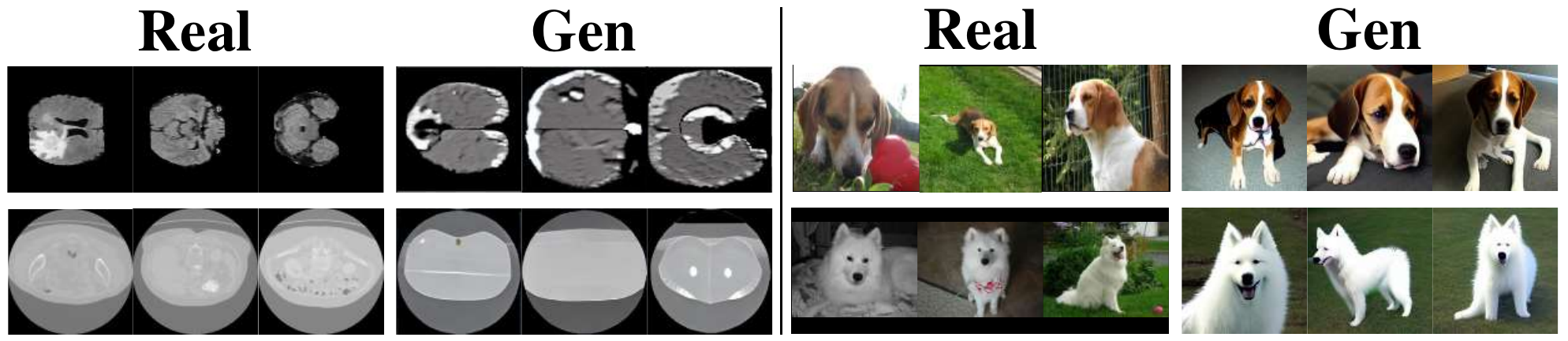}
    \caption{The most similar real-generated image pairs.}
    \label{fig-generalizability}
\end{figure}

We demonstrate the generalization capability of DDDR in the following two points:
1) We validated FCI's generative ability on widely used medical image datasets (LiTS \cite{bilic2023liver} and MSD \cite{antonelli2022medical}) and fine-grained classification datasets (Stanford Dogs \cite{dataset2011novel}). The results in Figure \ref{fig-generalizability} show that FCI can effectively generate data even when there are significant differences from the pretraining data.
2) The CIFAR-100 and TinyImageNet datasets we used were not used for pretraining the diffusion model \cite{schuhmann2021laion}. 

\section{Local Test Result}\label{sec_local_test}

% camera-ready
\begin{table}[h]
\tabcolsep=1.3mm
\centering 
\caption{Results of the comparative experiments on the Cifar-100 dataset. `T' indicates the task number. `Acc' denotes average accuracy, with higher values indicating better performance, and `FM' represents the forgetting measure, where lower values signify lesser forgetting of historical tasks. The best results are highlighted in bold.}
\label{tab-cifar}
\scalebox{0.7}{
\begin{tabular}{c|cccc|cccc}
\toprule
Data partition & \multicolumn{4}{c|}{IID}                                                        & \multicolumn{4}{c}{non-IID}                                                    \\ \midrule
Tasks          & \multicolumn{2}{c|}{T=5}                              & \multicolumn{2}{c|}{T=10} & \multicolumn{2}{c|}{T=5}                            & \multicolumn{2}{c}{T=10} \\ \midrule
Method         & Acc$(\uparrow)$& \multicolumn{1}{c|}{FM$(\downarrow)$}& Acc$(\uparrow)$& FM$(\downarrow)$& Acc$(\uparrow)$& \multicolumn{1}{c|}{FM$(\downarrow)$}& Acc$(\uparrow)$& FM$(\downarrow)$\\ \midrule
Finetune       & 17.33$\pm$0.18          & \multicolumn{1}{c|}{0.83$\pm$0.01}            & 9.03$\pm$0.18        & 0.88$\pm$0.01        & 16.47$\pm$1.12          & \multicolumn{1}{c|}{0.74$\pm$0.07}          & 8.58$\pm$0.58        & 0.77$\pm$0.05       \\
FedEWC         & 21.35$\pm$0.49          & \multicolumn{1}{c|}{0.69$\pm$0.01}            & 11.76$\pm$0.50       & 0.73$\pm$0.01        & 20.94$\pm$1.20          & \multicolumn{1}{c|}{0.61$\pm$0.05}          & 11.56$\pm$1.14       & 0.67$\pm$0.07       \\
Target         & 34.40$\pm$0.97          & \multicolumn{1}{c|}{0.48$\pm$0.01}            & 22.95$\pm$0.55       & 0.49$\pm$0.01        & 34.37$\pm$2.30          & \multicolumn{1}{c|}{0.48$\pm$0.04}          & 21.68$\pm$2.27       & 0.53$\pm$0.04       \\
MFCL           & 42.67$\pm$0.82          & \multicolumn{1}{c|}{0.37$\pm$0.01}            & 31.35$\pm$0.52       & 0.46$\pm$0.01        & 41.16$\pm$2.57          & \multicolumn{1}{c|}{0.33$\pm$0.03}          & 28.92$\pm$2.14       & 0.43$\pm$0.03       \\
Ours           & \textbf{51.04$\pm$0.83} & \multicolumn{1}{c|}{\textbf{0.29$\pm$0.01}}   & \textbf{43.45$\pm$0.76}   & \textbf{0.32$\pm$0.01}   & \textbf{48.45$\pm$3.56} & \multicolumn{1}{c|}{\textbf{0.26$\pm$0.04}} & \textbf{41.14$\pm$4.57}   & \textbf{0.30$\pm$0.04}  \\ \bottomrule
\end{tabular}
}
\end{table}

\begin{table}[h]
\tabcolsep=0.8mm
\centering 
\caption{Results of the comparative experiments on the Tiny-ImageNet dataset.}
\label{tab-tinyimagenet}
\scalebox{0.7}{
\begin{tabular}{c|cccc|cccc}
\toprule
Data partition & \multicolumn{4}{c|}{IID}                                               & \multicolumn{4}{c}{non-IID}                                           \\ \midrule
Tasks          & \multicolumn{2}{c|}{T=5}                   & \multicolumn{2}{c|}{T=10} & \multicolumn{2}{c|}{T=5}                   & \multicolumn{2}{c}{T=10} \\ \midrule
Method         & Acc$(\uparrow)$& \multicolumn{1}{c|}{FM$(\downarrow)$}& Acc$(\uparrow)$& FM$(\downarrow)$& Acc$(\uparrow)$& \multicolumn{1}{c|}{FM$(\downarrow)$}& Acc$(\uparrow)$& FM$(\downarrow)$\\ \midrule
Finetune       & 12.29$\pm$0.46     & \multicolumn{1}{c|}{0.60$\pm$0.01}      & 6.80$\pm$0.29        & 0.67$\pm$0.01        & 11.68$\pm$0.61     & \multicolumn{1}{c|}{0.52$\pm$0.04}      & 6.57$\pm$0.67        & 0.59$\pm$0.03       \\
FedEWC         & 13.27$\pm$0.45     & \multicolumn{1}{c|}{0.49$\pm$0.01}      & 8.22$\pm$0.30        & 0.56$\pm$0.01        & 12.55$\pm$0.70     & \multicolumn{1}{c|}{0.43$\pm$0.03}      & 7.67$\pm$0.90        & 0.50$\pm$0.03       \\
Target         & 17.56$\pm$0.49     & \multicolumn{1}{c|}{0.45$\pm$0.01}      & 12.53$\pm$0.43       & 0.49$\pm$0.01        & 17.88$\pm$0.85     & \multicolumn{1}{c|}{0.43$\pm$0.03}      & 11.31$\pm$0.90       & 0.47$\pm$0.03       \\
MFCL           & 15.11$\pm$0.47     & \multicolumn{1}{c|}{0.52$\pm$0.01}      & 10.13$\pm$0.48       & 0.54$\pm$0.01        & 13.31$\pm$1.18     & \multicolumn{1}{c|}{0.45$\pm$0.03}      & 8.57$\pm$0.45        & 0.49$\pm$0.02       \\
Ours           & \textbf{25.47$\pm$0.85} & \multicolumn{1}{c|}{\textbf{0.36$\pm$0.01}} & \textbf{19.01$\pm$0.67}   & \textbf{0.36$\pm$0.01}   & \textbf{23.97$\pm$1.26} & \multicolumn{1}{c|}{\textbf{0.34$\pm$0.03}} & \textbf{16.63$\pm$0.75}   & \textbf{0.32$\pm$0.04}  \\ \bottomrule
\end{tabular}
}
\end{table}

Our results presentation in the main text follows the mainstream work in the FCCL field \cite{qi2022better,zhang2023target,babakniya2024data}, calculating metrics on a global test set. However, to demonstrate performance variations across different clients, we also report the mean and standard deviation of metrics from multiple clients' independent tests. 
From the results in Tables \ref{tab-cifar} and \ref{tab-tinyimagenet}, we can draw the same conclusion as in the main text, namely that our method significantly outperforms the others.

\end{document}